\pdfoutput=1
\documentclass[11pt]{article}

\usepackage[final]{acl}

\usepackage{times}
\usepackage{latexsym}
\usepackage[dvipsnames]{xcolor}
\usepackage{amsmath,bm}
\usepackage{graphicx} 
\usepackage{cuted}
\usepackage{float}
\usepackage{setspace}

\usepackage{gb4e}
\noautomath
\usepackage[T1]{fontenc}
\usepackage{hyperref}
\usepackage{tcolorbox}
\usepackage{enumitem} % for customizing lists

\usepackage[utf8]{inputenc}

\usepackage{linguex}

\usepackage[labelfont=bf]{caption}

\usepackage{microtype}
\usepackage{dblfloatfix}

\usepackage{inconsolata}

\usepackage{graphicx}

\title{Can LLMs Detect Ambiguous Plural Reference? An Analysis of Split-Antecedent and Mereological Reference}

\author{Anh Dang\textsuperscript{1}, 
        Rick Nouwen\textsuperscript{1}, 
        Massimo Poesio\textsuperscript{1,2} \\
        \textsuperscript{1}Utrecht University,
        \textsuperscript{2}Queen Mary University of London
        \\
        \{t.t.a.dang, R.W.F.Nouwen, m.poesio\}@uu.nl}

\begin{document}
\maketitle
\begin{abstract}

Our goal is to study how LLMs represent and interpret plural reference in ambiguous and unambiguous contexts. We ask the following research questions: (1) Do LLMs exhibit human-like preferences in representing plural reference? (2) Are LLMs able to detect ambiguity in plural anaphoric expressions and identify possible referents? To address these questions, we design a set of experiments, examining pronoun production using next-token prediction tasks, pronoun interpretation, and ambiguity detection using different prompting strategies. We then assess how comparable LLMs are to humans in formulating and interpreting plural reference. We find that LLMs are sometimes aware of possible referents of ambiguous pronouns. However, they do not always follow human reference when choosing between interpretations, especially when the possible interpretation is not explicitly mentioned. In addition, they struggle to identify ambiguity without direct instruction. Our findings also reveal inconsistencies in the results across different types of experiments.

\end{abstract}

\section{Introduction}
Large language models (LLMs) have been shown to reach human-like competence in a wide range of natural language tasks \cite{touvron2023llama, brown2020language}. Communication between humans and AI systems is becoming more ubiquitous, given their remarkable language understanding and generation capacity. An important factor in effective communication is the ability to handle ambiguity in language. Although humans are very good at resolving ambiguity based on context and common sense knowledge, LLMs are designed with a strong focus on accuracy. In fact, they have been shown to struggle with capturing language ambiguity \cite{liu2023we, wildenburg2024pre, stengel2023zero, karamolegkou2025trick}. They are often unable to detect or generate ambiguous expressions. Since ambiguity is always present in communication, this inadequacy can lead to misunderstanding and the inability of LLMs to fully comprehend the use of language in context. 

\begin{figure}
    \centering
    \small{
    \includegraphics[width=1\linewidth]{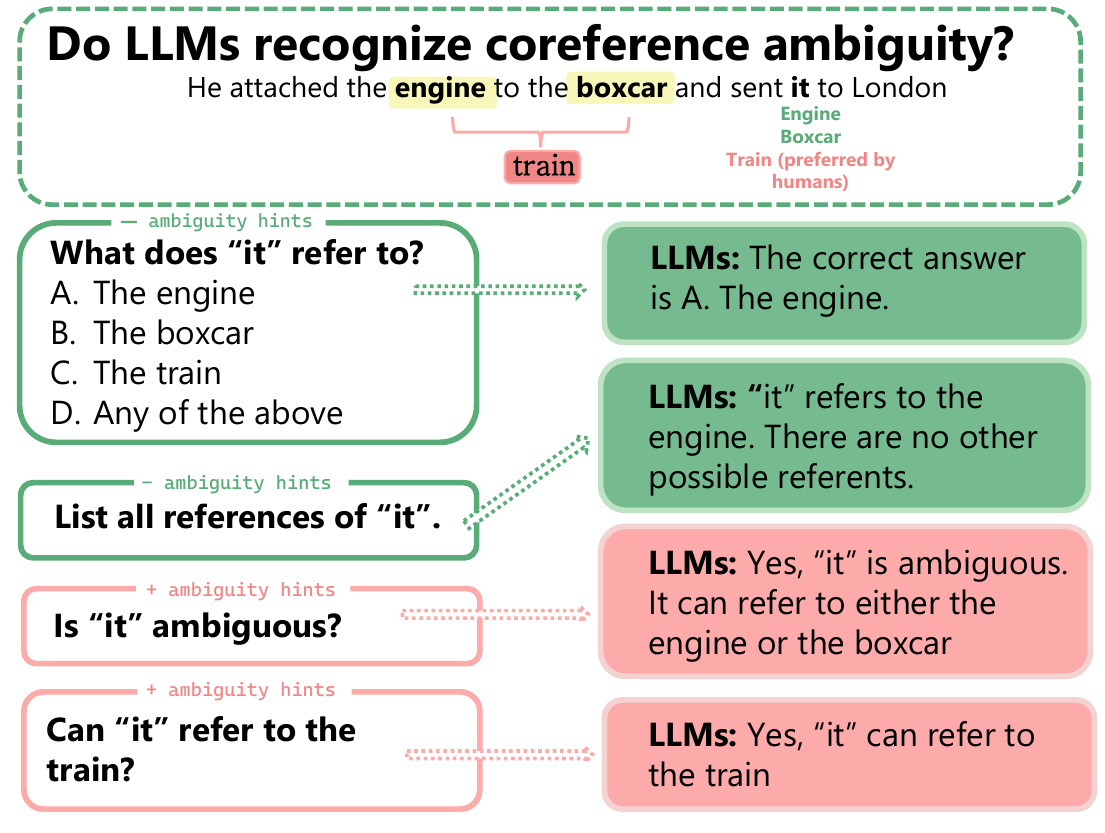}
    \caption{An illustration of the setup and results of our prompting experiments about how the extent to which LLMs can detect ambiguity in anaphoric expressions. Psycholinguistic studies show that in the ambiguous sentence above, humans prefer using \textit{it} to refer to combination of the \textit{engine} and the \textit{boxcar} or to one of the objects. We found that LLMs' responses vary depending on how much information about ambiguity is present in the prompt.}
    \label{fig:placeholder}
    }
\end{figure}

In this work, we focus on the ambiguity in coreference resolution. A lot of work has been done on building coreference resolution systems \cite{yu2020free, lee2017end, bohnet2023coreference}. More recent efforts to evaluate LLMs have shown that they can resolve coreference to some extent, yet it is still a challenging task \cite{brown2020language, pradhan2012conll, gan2024assessing}. In actual communication,  anaphoric expressions can sometimes be ambiguous. This means that it is not always clear to which referent the anaphoric expression refers. For example, when we say \textit{Jane looked at Mary and she smiled}, the pronoun \textit{she} possibly refers to Jane or Mary. Without further context, it is impossible to disambiguate the pronoun. 

Some research has been conducted on how LLMs process anaphoric expressions. However, most of the work focuses on singular reference and specific cases of ambiguity such as gender-ambiguous pronouns \cite{webster2018mind, emami2019knowref}. No research has been done specifically on the processing of plural reference. In this paper, our aim is to study how LLMs conceptualize plural reference in both ambiguous and unambiguous cases. We find plural reference particularly interesting because it is highly dependent on how humans conceptualize plurality. In psycholinguistic research, formulating plural expressions is a complex process because they introduce ambiguity to the discourse \cite{cokal2023anaphoric, koh2002resolution, moxey2012representing, moxey2004constraints, asher2003ambiguity, patson2014processing}. Our goal is to study \textbf{how LLMs produce, interpret, and represent plural reference in unambiguous and ambiguous contexts}.

We design a set of experiments, including next-token prediction and prompting, to compare how comparable LLMs are to humans in producing and interpreting ambiguous and unambiguous plural reference. We ask the following research questions.

%\vspace{0.2cm}
\paragraph{RQ1:} \textit{Do LLMs exhibit human-like preferences when representing plural reference?}

\paragraph{RQ2:} \textit{Are LLMs able to detect ambiguity in plural anaphoric expressions and identify possible referents?}

Our main findings are as follows: (1) LLMs follow human preference when choosing between interpretations in ambiguous cases where the possible referents are directly mentioned and struggle to do so when they are not present in the input; (2) LLMs fail to identify ambiguity without direct instruction.

\section{Background}

There is a body of work in psycholinguistics that shows that plural reference is more difficult to process and interpret than singular reference because the antecedents of plural anaphoric expressions are usually underspecified \cite{patson2014processing}, leaving more possibilities to interpret them. For example, in the pair of sentences below, the pronoun \textit{they} is more likely to refer to three people in 1(a) than 1(b). In 1(b), \textit{they} may also refer to the group of Tom and Dick.

\ex.
    \a. Tom, Dick, and Harry went to a restaurant last night. They ate a pizza.
    \b. Tom and Dick went to a restaurant with Harry last night. They ate a pizza.

 Here, we focus specifically on the cases where there is an interaction between singular and plural reference. These are cases where singular and plural expressions are both probable. \citet{moxey2012representing} find that the degree of overlap in roles between objects affects whether they are grouped as a plural entity. For example, in 
 sentences % MP sentence 
 2(a) and 2(b), \textit{waitress} is more related to \textit{waiter} than \textit{customer} in that they are more likely to perform the same action. As such, \textit{they} is more likely to be the continuation of 2(a) than 2(b). \\

\ex. 
  \a. The waiter and the waitress were in the restaurant. \_\_ (\textbf{He/She/They}) 
  \b. The waiter and the customer were in the restaurant. \_\_ (\textbf{He/She/They})

More recently, \citet{cokal2023anaphoric} investigate how humans interpret singular and plural pronouns in 
contexts %context 
where there is a whole-part relation between entities. 
When two singular objects are introduced to the discourse, and they are combined into a unified object, the new unified object is known as a \textit{mereological} object \cite{poesio2006underspecification}. It has been found that when more than one object is present, a plural expression is preferred and the use of a singular expression can cause processing difficulty \cite{moxey2004constraints, sanford1995notes, gordon1999processing}. However, there are cases where a singular reference may become salient as a new entity is created by combining the two antecedents.

\ex. 
  \a. The engineer attached the \textit{engine} to the \textit{boxcar} and sent \_\_ (\textbf{it/them})
  \b. The engineer detached the \textit{engine} from the \textit{boxcar} and sent \_\_ (\textbf{it/them})

Consider the two sentences above. In 3(a), through the verb \textit{attached}, the engine and the boxcar are combined, introducing a new singular object to the discourse. This new object is a \textit{train}, though it is never explicitly mentioned. Now, a singular or plural reading is both appropriate. The pronoun following the verb \textit{sent} can be \textit{it} or \textit{them}, with \textit{it} referring to one of the parts (that is, the engine or the boxcar) or the new object (that is, the train). In 3(b), however, the verb \textit{detached} prevents the formation of the mereological entity, as such, when \textit{it} is used, it can only refer to one of the objects. 
\textit{Them} is used to refer to the group of two objects. \citet{cokal2023anaphoric} found that humans prefer using \textit{it} more in 3(a) than 3(b). Moreover, they are more likely to consider the mereological object rather than the constituents to be the referent of \textit{it}. This suggests that when entities are joined, a new unified \textit{ group} entity is created, making \textit{it} a preferred antecedent for singular pronouns. In the following sections, we study how LLMs produce and interpret pronouns in these ambiguous cases.

\section{Ambiguity in LLMs}

Ambiguity is 
when an expression can be interpreted in different ways in a context.
% MP the variation in the interpretation of an expression. 
The use of everyday language is full of ambiguity, and it is shown that ambiguity indeed facilitates effective communication by minimizing effort \cite{piantadosi2012communicative}. With LLMs, recognizing ambiguity can be challenging, as they are not trained for this purpose and do not have access to pragmatic contexts. 

An emerging body of research has been done on how LLMs handle ambiguity in input. \citet{stengel2023zero} focus on semantic parsing. They prompt LLMs with ambiguous sentences with two possible interpretations and ask them to respond with the logical parses that correspond to the interpretations. They found that LLMs are unable to capture the possible interpretations in zero-shot contexts but perform much better in the few-shot contexts. \citet{sedova2024know} look at entity ambiguity. They show that while LLMs are aware of the possible readings, they often struggle with explicitly verbalizing the ambiguity. \citet{liu2023we} use a prompting-based approach to test the ability of LLMs in disambiguating, recognizing, and modeling different types of linguistic ambiguity. They found that even \texttt{GPT-4} still struggles with capturing ambiguity.

In the domain of coreference resolution, some studies have considered coreference to be one of the main sources of ambiguity \cite{wildenburg2024pre, kamath2024scope, liu2023we, gan2024assessing}. However, the types of ambiguity they cover are quite limited \cite{wildenburg2024pre, kamath2024scope, liu2023we} or strongly focus on unambiguous reference \cite{gan2024assessing}. The instances related to coreference account for a small number in their datasets. As such, in this work, we aim for a more extensive investigation of different types of plural reference ambiguity.

\section{Methods}

We use two types of experiments to study how LLMs process ambiguous plural references. For both experiments, we test a number of recent autoregressive decoder-only LLMs, namely \texttt{Qwen2.5-7B} \cite{team2024qwen2}, \texttt{Aya-Expanse-8B} \cite{dang2024aya} and \texttt{Llama-3.1-8B} \cite{touvron2023llama}. We evaluate the ability of LLMs to process ambiguous plural anaphoric expressions along the following dimensions. These capabilities are tested using a combination of prompting and next-token prediction tasks.

\paragraph{Production} is the ability to predict the correct pronoun for the referents. We test whether LLMs are aware of the pronouns for all possible antecedents. For example, in mereological contexts, are they able to identify that both singular (mereological object) and plural pronouns (constituents) are appropriate? 

\paragraph{Interpretation} means being able to directly specify the possible antecedents of the pronoun. Several prompting experiments will be used to evaluate this ability.

\paragraph{Detection} refers to the ability to identify whether the pronoun can have multiple possible antecedents. This ability can be demonstrated through the extent to which LLMs can provide meta-linguistic judgment of the ambiguity.

%\textbf{Experiment 1. Pronoun Prediction} We employed next-token prediction task to study the processing of plural reference. We provide the LLMs with the sentence prefix and make them predict the next token, which is most likely to be the pronoun. We then compare the probabilities of the target pronouns. For example, in the mereological case, we compare the probabilities of \textit{it} and \textit{them} in the mereological and non-mereological sentences. We then compare the probabilities of plural and singular pronouns between LLMs and humans. We use human data from aforementioned psycholinguistic studies. Below are the types of plural reference that we study along with examples.

%\textbf{Experiment 2. Ambiguity Detection} We use prompting techniques to test whether LLMs are able to detect and verbalize coreference ambiguity. We first prompt LLMs to directly answer whether the pronoun of the sentence is ambiguous. We then use a multiple-choice question-answering format to make the LLMs choose the referent for the pronoun. The goal of the task is to elicit models' interpretation of the pronoun's referent. 

\section{Production}

\subsection{Simple Plural References}

We first look at the case of simple plural reference as a baseline experiment, where the referents of the pronouns are completely unambiguous. When the direct object is singular, the pronoun referring to that object must also be singular. In the experiment, we asked the LLMs to predict the pronoun for both singular and plural referents.

\paragraph{Dataset} We use \texttt{GPT-4o} to generate the dataset \footnote{The GitHub repository consisting of the datasets and code is available at \url{https://github.com/dangthithaoanh/ambiguous-plural-reference} }. It consisted of 74 pairs of sentence prefixes. After the generation, we manually inspect each sentence pair to ensure that they meet the following requirements: (1) The direct object must be a countable object and (2) the second verb must be plausible with both the singular and plural versions of the direct object. Each pair differs in the number of the direct object, causing a difference in the choice of pronoun following the second verb. For each sentence prefix, there is only one correct pronoun. See the pair of sentences below for an example.

\ex. 
  \a. The Michelin chef prepared the dish and served \_\_
  \b. The Michelin chef prepared the dishes and served  \_\_

\paragraph{Results} Table ~\ref{sing-plur} shows the mean probability of singular and plural pronouns and the difference between them. As expected, we found that LLMs produce much higher probabilities for the correct pronouns for the sentence prefixes. This shows that, in simple unambiguous contexts, LLMs are fully capable of producing correct pronouns for the antecedents.

\begin{table}[t]
\centering
{\small
\begin{tabular}{lcccc}
\hline
\textbf{Model} & \textbf{Condition} & $\boldsymbol{P_{sing}}$ & $\boldsymbol{P_{plur}}$ & \textbf{Diff.} \\
\hline
\hline
 \texttt{Aya} & Singular & .336  & .009 & .326 \\
& Plural & .014  & .484 & -.470 \\
\hline
\texttt{Llama} & Singular & .352 & .009 & .343 \\
& Plural & .012 &  .467 & -.455 \\
\hline
\texttt{Qwen} & Singular & .576 & .007 & .568  \\
& Plural & .021 & .478  & -.456 \\

\hline

\end{tabular}
}
\caption{Results of Probability Prediction Task for Unambiguous Reference. \texttt{Diff.} is the probability difference between singular and plural pronouns ($P_{it} - P_{them}$), averaged across examples.}
\label{sing-plur}
\end{table}

\subsection{Split Antecedents Reference}

\citet{moxey2012representing} find that the degree of overlap between roles assigned to characters predicts the formation of a plural reference object. If LLMs exhibit a preference similar to that of humans, we would expect that the probabilities of the plural pronoun in 2(a) are higher than in 2(b), as shown in \citet{moxey2012representing}. Following the study, we include another condition in which the two characters are linked by the conjunction \textit{with} instead of \textit{and}. There are studies showing that the conjunction \textit{with} may decrease the preference for using plural pronouns \cite{albrecht1998accessing,sanford1990description}. The pair of sentences below shows the difference between the \textit{and} and \textit{with} conditions.

\ex. 
  \a. The waiter and the waitress were in the restaurant. \_\_
  \b. The waiter was in the restaurant with the waitress.  \_\_

\paragraph{Dataset} The dataset is adopted from the materials in \citet{moxey2012representing}. %Mp replaced , with .
Each condition consists of 24 sentence prefixes. The conditions include one with a high overlap between roles or spatial status. In addition to the overlap of the roles, there are also two other conditions where the conjunction linking the two roles is different, namely \textit{and} and \textit{with}. Using a \texttt{2x2} setting, this results in 4 conditions per example. See Table ~\ref{data} for an example of all conditions. %of the dataset. 

\paragraph{Metric} For each prefix, we extract and compare the probability difference between singular and plural pronouns. Possible singular pronouns include \textit{he} and \textit{she} ($P_{he/she}$). 
We 
consider %take into account 
the pronoun with 
the highest %the higher 
probability. 
Plural pronoun is \textit{they} ($P_{they}$). 

\begin{table}[htbp]
\centering
{\small
\begin{tabular}{lcccc}
\hline
\textbf{Model} &  \textbf{Condition} & $\boldsymbol{P_{sing}}$ & $\boldsymbol{P_{plur}}$ & \textbf{Diff.} \\
\hline
\hline

\texttt{Aya}  & High--And & .015  & .163 & -.148  \\
& High--With & .047 & .145  &  -.097  \\
& Low--And & .016 & .100 & -.084 \\
& Low--With & .047  & .102 & -.054 \\
\hline

 \texttt{Llama} & High--And & .013 & .129 & -.116 \\
& High--With & .052 & .153 & -.101  \\
& Low--And & .014 & .065 & -.051  \\
& Low--With &  .065 & .116 & -.051 \\

\hline

\texttt{Qwen}  & High--And & .003 & .151 &  -.147 \\
& High--With & .022 & .125  &  -.103 \\
& Low--And & .003  &  .075 &  -.072 \\
& Low--With &  .039  & .088 &  -.048 \\

\hline

\end{tabular}
}
\caption{Results of Probability Prediction Task for Split-antecedent Plural References. The \textit{High} and \textit{Low} conditions represent high or low overlap between roles. \textit{And} and \textit{With} are the conjunction linking the two roles. \texttt{Diff.} is the probability difference between the singular and plural pronoun ($P_{sing} - P_{plur}$), averaged across examples.}
\label{moxy}
\end{table}

\paragraph{Results} We compare the probabilities of singular and plural continuations between conditions. Table \ref{moxy} reports the mean probabilities between the targeted pronouns in all conditions for both LLMs. We run a two-way ANOVA with \texttt{role overlap} and \texttt{conjunction} as independent variables. The dependent variable is the probability difference between the singular and plural pronouns. 

We find that the main effect of \texttt{role overlap} was significant (\texttt{Aya-Expanse-8B}: \textit{F}(1, 23) = 35.04, \textit{p} <.001; \texttt{Qwen2.5-7B}:  \textit{F}(1, 23) = 16.7, \textit{p} <.001; \texttt{Llama-3.1-8B}: \textit{F}(1, 23) = 17.96, \textit{p} <.001). The main effect of \texttt{conjunction} was also not significant for \texttt{Aya-Expanse-8B}  (\textit{F}(1, 23) = 1.23, \textit{p} = .277) but was significant for \texttt{Qwen2.5-7B} (\textit{F}(1, 23) = 5.85, \textit{p} = .023) and \texttt{Llama-3.1-8B} (\textit{F}(1, 23) = 33.37, \textit{p} < .001). The interaction effect between \texttt{conjunction} and \texttt{role overlap} was not significant (\texttt{Aya-Expanse-8B}: \textit{F}(1, 23) = 1.23, \textit{p} = .277; \texttt{Qwen2.5-7B}: \textit{F}(1, 23) = .44, \textit{p} = .512; \texttt{Llama-3.1-8B}: \textit{F}(1, 23) = 1.54, \textit{p} = .226). This result is consistent with \citet{moxey2012representing}, suggesting that the degree of overlap of the roles between the antecedents affects how LLMs assign probabilities to singular and plural pronouns.

\subsection{Mereological Reference}

We report experiments on the production of singular and plural pronouns in mereological contexts, a type of ambiguous plural reference where two distinct objects 
also form %can also be considered 
a unified object.

\paragraph{Dataset} The dataset of this task consists of 44 pairs of mereological and non-mereological sentences. Half of the dataset is adapted from the material from \citet{cokal2023anaphoric}. We manually generate more pairs of sentences following the same template. Before running the main experiment, we want to ensure that the model is aware of the mereological object formed by attaching the two objects. We annotate the dataset with the names of the mereological objects and create another version of the dataset, in which the original sentence prefixes are appended with the token \textit{the}. Each sentence in the dataset for this experiment thus has the following template: 
\texttt{\{The engineer attached the engine to the goods wagon and sent the; engine; wagon; train\}}

We then have the LLMs predict the next token of the mereological sentences, where the mereological objects may have a higher chance of being observed. We then look at the top 15 tokens with the highest probabilities. If the mereological object is one of those tokens, then the sentence prefix is qualified. Otherwise, we remove the prefix from the dataset. This results in a total of 44 pairs of sentences.

\begin{table}[h]
\centering
{\small
\begin{tabular}{lcccc}
\hline
\textbf{Model} & \textbf{Type} & $\boldsymbol{P_{it}}$ & $\boldsymbol{P_{them}}$ & \textbf{Diff.} \\
\hline
\hline
 \texttt{Aya} & Mereo. & .336  & .009 & .326 \\
& Non-mereo. & .484 & .014 & .470 \\
\hline
\texttt{Llama} & Mereo. & .352 & .009 & .343 \\
& Non-mereo. & .467 & .012 & .455 \\

\hline
\texttt{Qwen} & Mereo. & .363  & .005 & .357 \\
& Non-mereo. & .507 & .012 & .494 \\

\hline

\end{tabular}
}
\caption{Results of Probability Prediction Task for Mereological Reference. The \texttt{Diff.} column represents the mean probability difference between \textit{it} and \textit{them ($P_{it} - P_{them} $)}.}
\label{mereo}
\end{table}

\subsubsection{Experiment 1: Pronoun Production} In this experiment, we test whether LLMs prefer singular pronouns in mereological contexts when the new objects are created. Through a next-word prediction task, we extract the probabilities of \textit{it} and \textit{them} in the mereological and non-mereological sentences. If the pronoun is \textit{it}, it refers to the individual object (that is, engine or boxcar) or the mereological object (i.e., train).

\paragraph{Metric} We compare the probability difference between \textit{it} and \textit{them} in the mereological ($D_{mereo}$) and non–mereological ($D_{N.mereo}$) sentences. The following formulas show how we calculate the difference.

\begin{center}
$ D_{mereo} = P_{it|mereo}-P_{them|mereo} $ \\
\noindent
$ D_{N.mereo} = P_{it|N.mereo}-P_{them|N.mereo} $
\end{center}

If the difference is larger in the mereological sentences than the non-mereological sentences, then the LLMs show a higher preference for \textit{it}, as found in \citet{cokal2023anaphoric}. 

\paragraph{Results} Table ~\ref{mereo} demonstrates how LLMs assign probabilities to \textit{it} and \textit{them} in mereological and non-mereological cases. It can be seen that LLMs prefer using \textit{it} in both contexts. We run paired-sample \textit{t}-tests, comparing $D_{mereo}$ and $  D_{N.mereo}$ across all pairs of sentences. The results show that the difference is significantly larger in non-mereological contexts (\texttt{Aya-Expanse-8B}:     \textit{t} = -4.78, \textit{p} < .001; \texttt{Llama-3.1-8B}: \textit{t} = -3.88, \textit{p} < .001; \texttt{Qwen2.5-7B}:     \textit{t} = -5.73, \textit{p} <.001). The statistical results show the opposite of what we expected. LLMs give higher probability to the singular pronoun when processing non-mereological sentences. It should be noted that this result does not reflect whether LLMs give a higher preference to the mereological object in mereological contexts because we do not yet know what \textit{it} refers to. As such, we run another experiment, in which we consider the probabilities of the objects themselves.

\subsubsection{Experiment 2: Object Production} After the first experiment, we observe that \textit{it} has very high probabilities in both contexts. However, while the pronoun \textit{them} clearly refers to the group of two objects, the antecedent of the pronoun \textit{it} can be highly ambiguous. It can refer to either the mereological object or one of the constituents. %Notably, 
We also observe that the probability of 
%article 
\textit{the} is quite high:
in most cases, the next most probable token is \textit{the}, which can be followed by the individual object or the mereological object. 
Thus, evidence for mereological object preference can also be demonstrated when the direct object of the verb is specified. As such, we conducted another experiment with \textit{the} after the original prefix. Adding \textit{the} forces the LLMs to specify the object to be referenced. 

\paragraph{Dataset} We use the same dataset from Experiment 1 and append \textit{the} to the sentence prefixes. See Table ~\ref{data} for a complete example of the dataset.

\paragraph{Metric}  We compare the probabilities of the individual objects with higher probabilities ($P_{ind.}$) and the mereological object ($P_{mereo.}$). See the formulas below for the detailed calculation of the metric.

\begin{center}
$ D_{mereo} = P_{ind.|mereo}-P_{mereo.|mereo} $ \\
\noindent
$ D_{N.mereo} = P_{ind.|N.mereo}-P_{mereo.|N.mereo} $
\end{center}

If the probability difference between the mereological objects and the constituents in the mereological cases ($ D_{mereo}$) is higher than that in the non-mereological cases ($ D_{N.mereo}$), then the LLMs show a greater preference to choose the mereological objects as the referent for the pronoun \textit{it} in mereological contexts. 

\begin{table}[h]
\centering
{\small
\begin{tabular}{lcccc}
\hline
\textbf{Model} & \textbf{Type} & $\boldsymbol{P_{ind.}}$ & $\boldsymbol{P_{mereo.}}$ & \textbf{Diff.} \\
\hline
\hline
 \texttt{Aya} & Mereo. & .076  & \textbf{.114} & -.038 \\
& Non-mereo. & .099 & .054 & .044 \\
\hline
\texttt{Llama} & Mereo. & .091 & .081 & .009  \\
& Non-mereo. & .134 & .048 & .085 \\
\hline
\texttt{Qwen} & Mereo. & .059 & \textbf{.191} & -.132  \\
& Non-mereo. & .206 & .045  &  .160 \\

\hline

\end{tabular}
}
\caption{Results of Probability Prediction Task for Object Production Experiment. \texttt{Diff.} is the mean probability difference between $P_{ind.}$ and $P_{mereo.}$. }
\label{ind-mereo}
\end{table}

\paragraph{Results} In Table ~\ref{ind-mereo}, we display the mean probability of individual objects, the mereological objects, as well as the difference between them. For \texttt{Aya-Expanse-8B} and \texttt{Qwen2.5-7B}, the mean probability of the mereological object is higher than its constituents. Paired-sample \textit{t}-tests show that $ D_{mereo}$ is significantly larger than $ D_{N.mereo}$ (\texttt{Aya-Expanse-8B}: \textit{t} = -2.845, \textit{p} = .006; \texttt{Llama-3.1-8B}: \textit{t} = -2.669, \textit{p} = .01 ; \texttt{Qwen2.5-7B}: \textit{t} = -5.825 , \textit{p} < .001). This result suggests that when directly choosing an object for the second verb (e.g. \textit{sent}), the LLMs do prefer the joined object more in the mereological contexts.

\begin{table*}[t]
\centering
\resizebox{\textwidth}{!}{%
\begin{tabular}{clcc}
\hline
\textbf{Reference Type} & \textbf{Sentence prefix} & \textbf{Expected pronoun} & \textbf{Metrics} \\
\hline
\hline
Simple Singular/Plural & The chef prepared the dish and served \_\_ & \colorbox{SkyBlue}{it}/them & $P_{it} - P_{them}$ \\
 & The chef prepared the dishes and served \_\_ & it/\colorbox{SkyBlue}{them} &  \\

\hline

Split-Antecedent & The waiter and the waitress were in the restaurant. \_\_ & \colorbox{Lavender}{he}/\colorbox{Lavender}{she}/\colorbox{Lavender}{they} &$P_{he/she} - P_{them}$ \\

\citet{moxey2004constraints, moxey2012representing} & The waiter was in the  restaurant with the waitress. \_\_ & & \\

& The waiter and the customer were in the  restaurant. \_\_ & & \\

& The waiter was in the restaurant with the customer. \_\_ & & \\

\hline

Mereological & The engineer attached the engine to the goods wagon and sent \_\_ & \colorbox{Lavender}{it}/\colorbox{Lavender}{them} & $D_{mereo} - D_{N.mereo}$ \\

\citet{cokal2023anaphoric} & The engineer detached the engine from the goods wagon and sent \_\_ &  &  \\

& The engineer attached the engine to the goods wagon and sent the \_\_ & \colorbox{Lavender}{mereo. obj.}/ \colorbox{Lavender}{ind obj}. & $D_{mereo.} - D_{N.mereo}$ \\

& The engineer detached the engine from the goods wagon and sent the \_\_ &  &  \\

\hline

\end{tabular}
}
\caption{Examples of the datasets used in all experiments. The types with expected pronouns in \colorbox{SkyBlue}{blue} mean that there can only be one possible pronoun and it is unambiguous. Those in \colorbox{Lavender}{pink} mean that multiple pronouns are possible.}
\label{data}
\end{table*}

\section{Interpretation}

While Section 5 aims at testing whether LLMs are able to produce correct pronouns in unambiguous and ambiguous contexts, in this Section we look at how LLMs interpret these pronouns. Do they interpret them differently in ambiguous contexts? Do they notice possible ambiguity? To do this, we use various prompting techniques to study the extent to which LLMs can detect ambiguity and generate possible referents for the pronouns.

Previous studies have found that while LLMs may be aware of the potential ambiguity in input, they are sometimes unable to transform this knowledge into a textual explanation \cite{stengel2023zero, karamolegkou2025trick}. Taking into account such a possibility, we use two prompting directions. In Prompt 1 (\textbf{P1}), we ask LLMs to 
provide all possible referents for a pronoun.
%choose the referent for the pronoun, providing all possible referents. 
In Prompt 2 (\textbf{P2}), 
LLMs have to  choose the referent for the pronoun among those in a supplied list.
%the possible referents list all potential referents for the pronoun.
Some of the listed referents suggest ambiguity,
%We next add some hints of ambiguity 
to see to what extent the LLMs can make use of them to detect ambiguity. 
In Prompt 3 (\textbf{P3}), we ask LLMs if the mereological object can be the referent. 
Finally, we ask if the LLMs think that the pronoun \textit{it} is ambiguous (\textbf{P4}). 
All of our prompts are few-shot.
See Appendix ~\ref{sec:appendix} for details about our prompt designs and how we code LLMs' responses. 
%All of our prompts are few-shot. 

For baseline comparison, we also prompt LLMs on the simple reference dataset. The experiments are run on two LLMs, namely \texttt{Aya-Expanse-8B} and \texttt{Qwen2.5-7B}.

\subsection{Reference Identification} 

\paragraph{Unambiguous Reference} As a baseline, we first test whether LLMs can choose the correct antecedent for simple unambiguous reference using the dataset described in Section 5.1. The results are not surprising. The LLMs reach 100\% accuracy. This shows that in unambiguous contexts, LLMs are fully able to choose the correct antecedents for pronouns.

\paragraph{Ambiguous Reference} We report the results of \textbf{P1} and \textbf{P2} for the mereological dataset. When being asked to list all possible referents for \textit{it}, the LLMs show somewhat different patterns. \texttt{Aya-Expanse-8B} consistently chooses one of the constituents as the only referent in both mereological and non-mereological contexts. \texttt{Qwen2.5-7B} is better at recognizing that the pronoun is ambiguous and that both objects can be the referent (50\%). However, both rarely mention the mereological object as a possible referent (11.3\%). 

Figure ~\ref{fig:plots} (top, left)  shows the percentage of each answer choice in mereological and non-mereological contexts. When LLMs are forced to choose an antecedent for 
%the 
pronoun \textit{it}, they generally prefer to link 
%the pronoun 
\textit{it} to the first object. Notably, for \texttt{Qwen2.5-7B}, the number of times the LLMs choose the first object is much higher 
%than 
in the non-mereological cases than in the mereological cases. However, \texttt{Aya-Expanse-8B} gives nearly equal preference for the first object and the mereological object. The preference for the second object is quite similar in both cases. As a result, the percentage of choosing the mereological object is higher in the mereological cases (\texttt{Aya-Expanse-8B}: 37.3\%, \texttt{Qwen2.5-7B}: 22\%) than the non-mereological cases (22.8\%). This shows that LLMs give higher preference to the mereological object in mereological contexts. However, they seem to be more likely to link the pronoun \textit{it} to the constituent in both cases. 

Another answer choice that we are also interested in is \textit{D. any of the above objects}. If the LLMs choose \textit{D}, they think that all objects are equally likely to be the referent for the pronoun \textit{it}. In fact, if the models understand that \textit{it} is ambiguous, this option has to be the most-chosen one. The low percentage of this option shows that, when being asked, the LLMs tend to specify an object as a referent instead of accepting the ambiguity. This shows that when the LLMs are instructed towards choosing a single referent, they often fail to identify the ambiguity that is present. 

A surprising result is that \texttt{Aya-Expanse-8B} chooses \textit{it} as referring to the mereological object quite often in the non-mereological prompts (21.3\%), where the mereological object is not available as a possible antecedent. 
This can be 
an %because of the 
effect of word frequency, such that the mereological object is more frequent or more likely to be preceded by the main verb.

\subsection{Ambiguity Detection}

\begin{figure*}[t]  % the [t] option places it at the top of the page
    \centering
    \includegraphics[width=\linewidth]{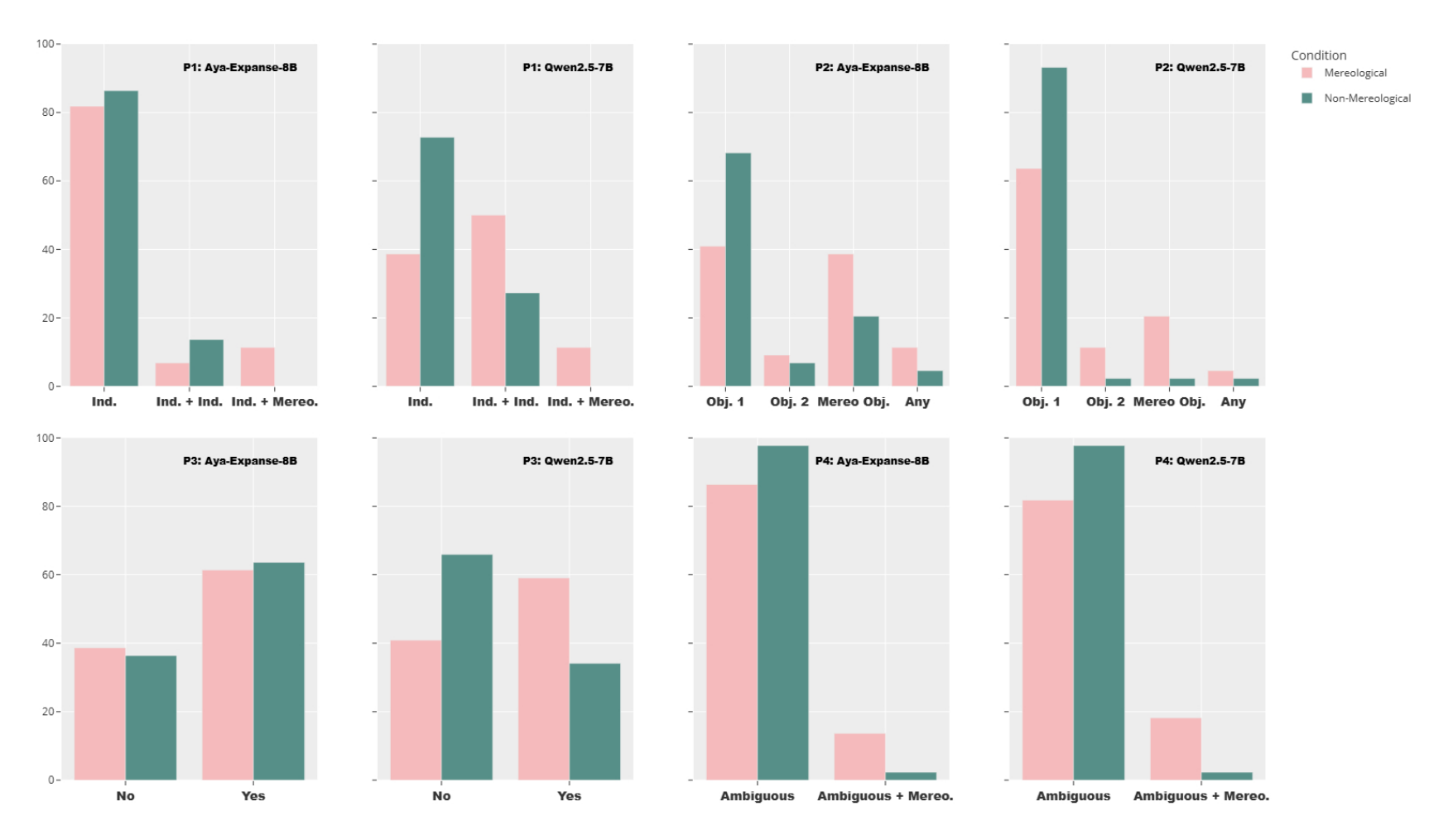}
    \caption{Results of prompting experiments with four different prompts (P1, P2, P3, and P4). The \textit{x}-axis displays different types of LLMs' responses. In \textbf{P1}, \texttt{Ind.} is the one of the constituent. \texttt{Ind. + Mereo.} means that they listed both the constituents and the mereological object. In \textbf{P2}, the LLMs choose between the constituents (\texttt{Obj.1} and \texttt{Obj. 2}, the mereological object (\texttt{Mereo. Obj.} or any of the mentioned objects (\texttt{Any}). In \textbf{P3}, they answered \texttt{Yes} or \texttt{No} to whether the mereological object can be the referent. In \textbf{P4}, the LLMs all identified \textit{it} as ambiguous. The \texttt{Ambiguous + Mereo.} column shows when the LLMs mention the mereological object as a possible referent.  }
    \label{fig:plots}
\end{figure*}

We now examine the results of \textbf{P3} and \textbf{P4} where we test the extent to which LLMs can detect ambiguity given direct mention. Recall that in \textbf{P4}, we ask the LLMs if \textit{it} is ambiguous. We also run this experiment on unambiguous sentences. We find that in 97.7\% of the cases, they correctly identify the pronoun as unambiguous.

Figure ~\ref{fig:plots} (bottom, right) shows the percentage of \textit{Yes} and \textit{No} responses. We found that LLMs consistently identify the pronoun \textit{it} as an ambiguous one (100\%) in both mereological and non-mereological
cases. This is rather surprising given that in the \textbf{P2} they almost never choose
the option that indicates ambiguity (i.e. any of the
above objects). In addition, in the explanations of
LLMs, they also generate possible antecedents of \textit{it}. In
13.6\% (\texttt{Aya-Expanse-8B}) and 18.2\% (\texttt{Qwen2.5-7B}) of the mereological prompts, they include the mereological object as a possible antecedent,
compared to 2.3\% in the non-mereological prompts.

%It can also be seen that both LLMs behave similarly. They always identify the pronoun as ambiguous. In their explanation, they list the two constituents as possible referents. Yet they also rarely mention the mereological object one of the possible candidate (\texttt{Aya-Expanse-8B}: 13.6\%, \texttt{Qwen2.5-7B}: 18.2\%). 

In \textbf{P3}, we ask the LLMs if the mereological object
can be the referent of \textit{it}. The percentage of \textit{Yes} in the mereological cases is moderate for both LLMs (\texttt{Aya-Expanse-8B}: 61.4\%, \texttt{Qwen2.5-7B}: 59.1\%).

\begin{tcolorbox}[colback=white!5!white,colframe=black!75!black,title=\small{SAMPLE RESPONSE - AYA-EXPANSE-8B}]

\small{<USER> Consider the referent of the pronoun 'it' in the following sentence: \\
The hardworking, experienced autoworker removed the suspension from the chassis and sent it \_\_. \\
Can 'it' refer to the car?} \\

<ASSISTANT> Yes, in the given sentence, the pronoun 'it' can refer to the car. The sentence is discussing an action performed on a car, specifically the removal of the suspension from the chassis. Since 'it' is used to refer to the object that is being acted upon, it logically follows that 'it' refers to the car.

\end{tcolorbox}

Notably, these  percentages are much higher than those in \textbf{P1} and \textbf{P2}. Yet, on the other hand, the difference between mereological and non-mereological contexts is quite small. Especially for \texttt{Aya-Expanse-8B}, the results are similar in the non-mereological cases, which is counterintuitive because the mereological object should not be available as a referent given the detaching action. The text above is an example response where \texttt{Aya-Expanse-8B} thinks that \textit{it} can refer to the mereological object in the non-mereological sentence. When the \textit{car} (mereological object) is mentioned in the prompt, the model believes that the suspension and the chassis are parts of a car. When the autoworker removes one of the parts (e.g., the suspension), the remaining is still a car. However, this line of reasoning does not appear in \textbf{P4}, when the mereological object is not mentioned in the non-mereological prompt. The model simply responds that both constituents can be the referent. This suggests a crucial difference in LLMs' preference when a potential referent is mentioned in the prompt or not. 

%Recall that we also observe this pattern in \textbf{P2}. We attribute this result to the frequency of the mereological objects compared to their constituents. 

%This shows that LLMs sometimes identify all possible antecedents for the pronoun it, yet this ability is still very limited given the low proportion.

\section{Discussion}

In the above experiments, we assess LLMs' ability to identify different types of coreference ambiguity using probability-based and prompting-based approaches. In the split-antecedent references, LLMs show a strong alignment with human preference \cite{moxey2004constraints, moxey2012representing}. When the roles assigned to the antecedents indicate that they may be involved in the same action, they are more likely to be referenced as a plural entity. However, in mereological contexts, the results are more complex. On the one hand, LLMs show preference for the mereological object following the verb, as shown in the \textit{Object Production} experiment. On the other hand, the \textit{Interpretation} experiment shows the lack of self-consistency of LLMs between prompting strategies and hints about ambiguity and the mereological objects. Without hints of ambiguity, they almost never recognize the ambiguity when being prompted with \textbf{P1} and \textbf{P2}, although these prompts do not exclude the possibility that the pronoun is ambiguous. They also do not explicitly mention the mereological object as a referent except when it is provided in the prompt. Generally, LLMs prefer to link \textit{it} to the first constituent. 

However, in \textbf{P2} and \textbf{P3}, where the mereological objects are presented in the prompts, they are chosen or listed much more frequently (40\%--60\%). This leads to the possibility that LLMs may exhibit this psycholinguistic phenomenon when direct instruction is provided. This hypothesis is complemented by the results of the \textit{Object Production} experiment where some LLMs (\texttt{Aya-Expanse-8B and Qwen2.5-7B}) give higher probabilities to mereological objects in mereological contexts. This possibility aligns with \citet{stengel2023zero} and \citet{karamolegkou2025trick}, who find that LLMs are better at producing different interpretations in few-shot contexts. Taken together, these results may imply that LLMs may not be able to apply their knowledge of ambiguity without specific instruction. In other words, they sometimes know the possible interpretations, but struggle to consistently show them in different types of instruction. This may be the case, as LLMs successfully detect ambiguity between the constituents in \textbf{P4}. However, the results of \textbf{P3}, where the difference between mereological and non-mereological contexts is not considerable, the mention of the mereological objects may not mean that the LLMs actually recognize the joined object. As mentioned before, the effect of frequency may be present. This suggests that the LLMs may be able to detect potential referents when they are mentioned in the input but fail to take into account the mereological object, a more complex referent. This suggests that while LLMs broadly capture the basic findings about plural pronoun preferential interpretation from classic papers such as \cite{moxey2004constraints, moxey2012representing}, their preferences differ from those captured in more recent literature such as \citet{cokal2023anaphoric}.

We also observe that, while the LLMs perform quite similarly in the next-word prediction task, they exhibit different patterns in prompting tasks. \texttt{Qwen2.5-7B} seems to align more with humans. The model indeed chooses the mereological objects more in mereological contexts.

%Taken together, these results show that in mereological context where humans prefer the joined object as the referent, (1) the LLMs are not able to detect ambiguity in coreference unless they are exposed to hints of ambiguity in the instructions and (2) they 

%Our results in the \textit{Ambiguity Detection} experiment also reveal that while LLMs are rarely able to choose the option indicating ambiguity, they are often aware of possible interpretations when prompted with a direct mention of the presence of ambiguity. 

\section{Conclusion}

In this paper, we conduct a set of experiments to investigate whether LLMs are able to capture different interpretations of ambiguous plural reference and whether their representations of reference are comparable to how humans do in psycholinguistic literature. Our mixed findings contribute to a larger picture of the degree to which LLMs possess knowledge of ambiguity and how this knowledge is used.

\section{Limitations}

Our study provides new insights into how LLMs process text with different types of coreference ambiguity. Yet it has a few limitations that need to be noted. First, in the mereological cases, we think that the frequency of the constituents and the mereological object affect its likelihood to be chosen by the LLMs in both the next-token prediction and prompt tasks. For example, in the sentence \textit{The shoemaker stripped the leather from the sole and inspected it \_\_}, the mereological object \textit{shoe} is much more common than the \textit{leather} or the \textit{sole}. Although we try to select the verb that can be done on both constituents and the mereological objects, this effect of frequency is likely to be present, leading the LLMs to choose the mereological object as the referent without knowledge of the mereological relation. Secondly, the LLMs that we tested are rather small (around 7-8B). Investigations into larger LLMs are important as they may exhibit higher competency in interpreting ambiguous pronouns. In addition, we rely on prompting results, which may not always reflect the representation and reasoning capability of the LLMs. Future research may use recent mechanistic interpretability approaches for a more thorough and dimension-wise investigation of how LLMs represent ambiguous references \cite{syed2024attribution, vig2020investigating, hanna2025incremental}. Such methods involve interventions on the LLMs' hidden activations, which potentially uncover the process underneath the LLMs' generation.

In addition, as pointed out by the reviewers, we may be quite strict in evaluating LLMs' ability to provide metalinguistic judgements about ambiguity because even humans can still have difficulty identifying it. We did not conduct the prompting experiments on humans, so we do not know their preferences. We acknowledge this as one of the limitations of the paper.

\section{Ethical Considerations}

The study uses only publicly available linguistic stimuli from previous research and synthetic data generated by LLMs. It does not include any data from human participants.

\section{Acknowledgments}

We would like to thank three anonymous reviewers for their insightful comments. This project is funded by the Dutch Research Council (NWO) through the AiNed Fellowship Grant (Dealing with Meaning Variation, NGF.1607.22.002) to Massimo Poesio. 

%\bibliography{cus

\begin{thebibliography}{32}
\providecommand{\natexlab}[1]{#1}

\bibitem[{Albrecht and Clifton(1998)}]{albrecht1998accessing}
Jason~E Albrecht and Charles Clifton. 1998.
\newblock Accessing singular antecedents in conjoined phrases.
\newblock \emph{Memory \& cognition}, 26(3):599--610.

\bibitem[{Asher and Wang(2003)}]{asher2003ambiguity}
Nicholas Asher and Linton Wang. 2003.
\newblock Ambiguity and anaphora with plurals in discourse.
\newblock In \emph{Semantics and Linguistic Theory}, pages 19--36.

\bibitem[{Bohnet et~al.(2023)Bohnet, Alberti, and Collins}]{bohnet2023coreference}
Bernd Bohnet, Chris Alberti, and Michael Collins. 2023.
\newblock Coreference resolution through a seq2seq transition-based system.
\newblock \emph{Transactions of the Association for Computational Linguistics}, 11:212--226.

\bibitem[{Brown et~al.(2020)Brown, Mann, Ryder, Subbiah, Kaplan, Dhariwal, Neelakantan, Shyam, Sastry, Askell et~al.}]{brown2020language}
Tom Brown, Benjamin Mann, Nick Ryder, Melanie Subbiah, Jared~D Kaplan, Prafulla Dhariwal, Arvind Neelakantan, Pranav Shyam, Girish Sastry, Amanda Askell, and 1 others. 2020.
\newblock Language models are few-shot learners.
\newblock \emph{Advances in neural information processing systems}, 33:1877--1901.

\bibitem[{Cokal et~al.(2023)Cokal, Filik, Sturt, and Poesio}]{cokal2023anaphoric}
Derya Cokal, Ruth Filik, Patrick Sturt, and Massimo Poesio. 2023.
\newblock Anaphoric reference to mereological entities.
\newblock \emph{Discourse Processes}, 60(3):202--223.

\bibitem[{Dang et~al.(2024)Dang, Singh, D'souza, Ahmadian, Salamanca, Smith, Peppin, Hong, Govindassamy, Zhao et~al.}]{dang2024aya}
John Dang, Shivalika Singh, Daniel D'souza, Arash Ahmadian, Alejandro Salamanca, Madeline Smith, Aidan Peppin, Sungjin Hong, Manoj Govindassamy, Terrence Zhao, and 1 others. 2024.
\newblock Aya expanse: Combining research breakthroughs for a new multilingual frontier.
\newblock \emph{arXiv preprint arXiv:2412.04261}.

\bibitem[{Emami et~al.(2019)Emami, Trichelair, Trischler, Suleman, Schulz, and Cheung}]{emami2019knowref}
Ali Emami, Paul Trichelair, Adam Trischler, Kaheer Suleman, Hannes Schulz, and Jackie Chi~Kit Cheung. 2019.
\newblock The knowref coreference corpus: Removing gender and number cues for difficult pronominal anaphora resolution.
\newblock In \emph{Proceedings of the 57th Annual Meeting of the Association for Computational Linguistics}, pages 3952--3961.

\bibitem[{Gan et~al.(2024)Gan, Poesio, and Yu}]{gan2024assessing}
Yujian Gan, Massimo Poesio, and Juntao Yu. 2024.
\newblock Assessing the capabilities of large language models in coreference: An evaluation.
\newblock In \emph{Proceedings of the 2024 Joint International Conference on Computational Linguistics, Language Resources and Evaluation (LREC-COLING 2024)}, pages 1645--1665.

\bibitem[{Gordon et~al.(1999)Gordon, Hendrick, Ledoux, and Yang}]{gordon1999processing}
Peter~C Gordon, Randall Hendrick, Kerry Ledoux, and Chin~Lung Yang. 1999.
\newblock Processing of reference and the structure of language: An analysis of complex noun phrases.
\newblock \emph{Language and Cognitive Processes}, 14(4):353--379.

\bibitem[{Hanna and Mueller(2025)}]{hanna2025incremental}
Michael Hanna and Aaron Mueller. 2025.
\newblock Incremental sentence processing mechanisms in autoregressive transformer language models.
\newblock In \emph{Proceedings of the 2025 Conference of the Nations of the Americas Chapter of the Association for Computational Linguistics: Human Language Technologies (Volume 1: Long Papers)}, pages 3181--3203.

\bibitem[{Kamath et~al.(2024)Kamath, Schuster, Vajjala, and Reddy}]{kamath2024scope}
Gaurav Kamath, Sebastian Schuster, Sowmya Vajjala, and Siva Reddy. 2024.
\newblock Scope ambiguities in large language models.
\newblock \emph{Transactions of the Association for Computational Linguistics}, 12:738--754.

\bibitem[{Karamolegkou et~al.(2025)Karamolegkou, Eberle, Rust, Kauf, and S{\o}gaard}]{karamolegkou2025trick}
Antonia Karamolegkou, Oliver Eberle, Phillip Rust, Carina Kauf, and Anders S{\o}gaard. 2025.
\newblock Trick or neat: Adversarial ambiguity and language model evaluation.
\newblock \emph{arXiv preprint arXiv:2506.01205}.

\bibitem[{Koh and Clifton~Jr(2002)}]{koh2002resolution}
Sungryong Koh and Charles Clifton~Jr. 2002.
\newblock Resolution of the antecedent of a plural pronoun: Ontological categories and predicate symmetry.
\newblock \emph{Journal of Memory and Language}, 46(4):830--844.

\bibitem[{Lee et~al.(2017)Lee, He, Lewis, and Zettlemoyer}]{lee2017end}
Kenton Lee, Luheng He, Mike Lewis, and Luke Zettlemoyer. 2017.
\newblock End-to-end neural coreference resolution.
\newblock In \emph{Proceedings of the 2017 Conference on Empirical Methods in Natural Language Processing}, pages 188--197.

\bibitem[{Liu et~al.(2023)Liu, Wu, Michael, Suhr, West, Koller, Swayamdipta, Smith, and Choi}]{liu2023we}
Alisa Liu, Zhaofeng Wu, Julian Michael, Alane Suhr, Peter West, Alexander Koller, Swabha Swayamdipta, Noah~A Smith, and Yejin Choi. 2023.
\newblock We’re afraid language models aren’t modeling ambiguity.
\newblock In \emph{Proceedings of the 2023 Conference on Empirical Methods in Natural Language Processing}, pages 790--807.

\bibitem[{Moxey et~al.(2004)Moxey, Sanford, Sturt, and Morrow}]{moxey2004constraints}
Linda~M Moxey, Anthony~J Sanford, Patrick Sturt, and Lorna~I Morrow. 2004.
\newblock Constraints on the formation of plural reference objects: The influence of role, conjunction, and type of description.
\newblock \emph{Journal of Memory and Language}, 51(3):346--364.

\bibitem[{Moxey et~al.(2012)Moxey, Sanford, and Tonks}]{moxey2012representing}
Linda~M Moxey, Anthony~J Sanford, and Karen Tonks. 2012.
\newblock Representing characters in a scenario: What makes two individuals a set?
\newblock \emph{Language and cognitive processes}, 27(9):1405--1424.

\bibitem[{Patson(2014)}]{patson2014processing}
Nikole~D Patson. 2014.
\newblock The processing of plural expressions.
\newblock \emph{Language and Linguistics Compass}, 8(8):319--329.

\bibitem[{Piantadosi et~al.(2012)Piantadosi, Tily, and Gibson}]{piantadosi2012communicative}
Steven~T Piantadosi, Harry Tily, and Edward Gibson. 2012.
\newblock The communicative function of ambiguity in language.
\newblock \emph{Cognition}, 122(3):280--291.

\bibitem[{Poesio et~al.(2006)Poesio, Sturt, Artstein, and Filik}]{poesio2006underspecification}
Massimo Poesio, Patrick Sturt, Ron Artstein, and Ruth Filik. 2006.
\newblock Underspecification and anaphora: Theoretical issues and preliminary evidence.
\newblock \emph{Discourse processes}, 42(2):157--175.

\bibitem[{Pradhan et~al.(2012)Pradhan, Moschitti, Xue, Uryupina, and Zhang}]{pradhan2012conll}
Sameer Pradhan, Alessandro Moschitti, Nianwen Xue, Olga Uryupina, and Yuchen Zhang. 2012.
\newblock Conll-2012 shared task: Modeling multilingual unrestricted coreference in ontonotes.
\newblock In \emph{Joint conference on EMNLP and CoNLL-shared task}, pages 1--40.

\bibitem[{Sanford and Lockhart(1990)}]{sanford1990description}
Anthony~J Sanford and F~Lockhart. 1990.
\newblock Description types and method of conjoining as factors influencing plural anaphora: A continuation study of focus.
\newblock \emph{Journal of Semantics}, 7(4):365--378.

\bibitem[{Sanford and Moxey(1995)}]{sanford1995notes}
Anthony~J Sanford and Linda~M Moxey. 1995.
\newblock Notes on plural reference and the scenario-mapping principle in comprehension.
\newblock \emph{Focus and coherence in discourse processing}, pages 18--34.

\bibitem[{Sedova et~al.(2024)Sedova, Litschko, Frassinelli, Roth, and Plank}]{sedova2024know}
Anastasiia Sedova, Robert Litschko, Diego Frassinelli, Benjamin Roth, and Barbara Plank. 2024.
\newblock To know or not to know? analyzing self-consistency of large language models under ambiguity.
\newblock In \emph{Findings of the Association for Computational Linguistics: EMNLP 2024}, pages 17203--17217.

\bibitem[{Stengel-Eskin et~al.(2023)Stengel-Eskin, Rawlins, and Van~Durme}]{stengel2023zero}
Elias Stengel-Eskin, Kyle Rawlins, and Benjamin Van~Durme. 2023.
\newblock Zero and few-shot semantic parsing with ambiguous inputs.
\newblock \emph{arXiv preprint arXiv:2306.00824}.

\bibitem[{Syed et~al.(2024)Syed, Rager, and Conmy}]{syed2024attribution}
Aaquib Syed, Can Rager, and Arthur Conmy. 2024.
\newblock Attribution patching outperforms automated circuit discovery.
\newblock In \emph{Proceedings of the 7th BlackboxNLP Workshop: Analyzing and Interpreting Neural Networks for NLP}, pages 407--416.

\bibitem[{Team(2024)}]{team2024qwen2}
Qwen Team. 2024.
\newblock Qwen2 technical report.
\newblock \emph{arXiv preprint arXiv:2407.10671}.

\bibitem[{Touvron et~al.(2023)Touvron, Lavril, Izacard, Martinet, Lachaux, Lacroix, Rozi{\`e}re, Goyal, Hambro, Azhar et~al.}]{touvron2023llama}
Hugo Touvron, Thibaut Lavril, Gautier Izacard, Xavier Martinet, Marie-Anne Lachaux, Timoth{\'e}e Lacroix, Baptiste Rozi{\`e}re, Naman Goyal, Eric Hambro, Faisal Azhar, and 1 others. 2023.
\newblock Llama: Open and efficient foundation language models.
\newblock \emph{arXiv preprint arXiv:2302.13971}.

\bibitem[{Vig et~al.(2020)Vig, Gehrmann, Belinkov, Qian, Nevo, Singer, and Shieber}]{vig2020investigating}
Jesse Vig, Sebastian Gehrmann, Yonatan Belinkov, Sharon Qian, Daniel Nevo, Yaron Singer, and Stuart Shieber. 2020.
\newblock Investigating gender bias in language models using causal mediation analysis.
\newblock \emph{Advances in neural information processing systems}, 33:12388--12401.

\bibitem[{Webster et~al.(2018)Webster, Recasens, Axelrod, and Baldridge}]{webster2018mind}
Kellie Webster, Marta Recasens, Vera Axelrod, and Jason Baldridge. 2018.
\newblock Mind the gap: A balanced corpus of gendered ambiguous pronouns.
\newblock \emph{Transactions of the Association for Computational Linguistics}, 6:605--617.

\bibitem[{Wildenburg et~al.(2024)Wildenburg, Hanna, and Pezzelle}]{wildenburg2024pre}
Frank Wildenburg, Michael Hanna, and Sandro Pezzelle. 2024.
\newblock Do pre-trained language models detect and understand semantic underspecification? ask the dust!
\newblock In \emph{Findings of the Association for Computational Linguistics ACL 2024}, pages 9598--9613.

\bibitem[{Yu et~al.(2020)Yu, Moosavi, Paun, and Poesio}]{yu2020free}
Juntao Yu, Nafise~Sadat Moosavi, Silviu Paun, and Massimo Poesio. 2020.
\newblock Free the plural: Unrestricted split-antecedent anaphora resolution.
\newblock In \emph{Proceedings of the 28th International Conference on Computational Linguistics}, pages 6113--6125.

\end{thebibliography}

\appendix

\section{Prompt Construction}
\label{sec:appendix}
\subsection{Prompt without Explicit Hints of Ambiguity}

\subsubsection{Prompt 1: Open-ended Questions}

In the first prompt, we ask the LLMs to generate all possible antecedents for \textit{it}. The goal is to see whether they have a deterministic choice or also consider other possible candidates.

\begin{tcolorbox}[colback=white!5!white,colframe=black!75!black,title=Prompt template 1]

\texttt{\small <USER> Consider the pronoun "it" in the following sentence prefix. What is the referent of "it"? List all possible referents if there are any. \\
Sentence: <sentence>} \\

\small \texttt{<ASSISTANT>}

\end{tcolorbox}

See the excerpt below for an example prompt and the LLM’s response.

\begin{quote}
<USER> Consider the referent of the pronoun 'it' in the following sentence: \\
Sentence: The bike rider clipped the wheel to the frame and checked it \_\_ \\
What is the referent of 'it'? List all possible referents if any.

<ASSISTANT> In the sentence "The bike rider clipped the wheel to the frame and checked it," the pronoun "it" likely refers to either "the wheel" or "the frame." \\
Given the sequence of actions described, the most logical interpretation would be that the bike rider has just clipped the wheel to the frame and then proceeded to check the frame for proper attachment or alignment. However, without additional context, there's also a possibility that "it" could refer to the wheel again, checking to ensure that the wheel has been securely attached to the frame. \\
Therefore, the possible referents for "it" are: \\
1. The wheel \\
2. The frame \\
\end{quote}

We divide LLMs' responses into two categories. If the LLMs only list one of the constituents, their responses are coded as \texttt{Ind.}. If they choose between the two constituents, these are coded as \texttt{Ind. + Ind.} In case they mention both constituents and the mereological object, responses are coded as \texttt{Ind. + Mereo.}

\subsubsection{Prompt 2: Multiple-choice Questions}

We append the expected pronoun to the sentence prefixes and ask the model to choose from the three objects using a multiple-choice format. To control for the effect of prompt difference, we use two types of prompts. In the multiple-choice question, there are two other options (\textit{D. any of the above objects}), which imply that the pronoun is ambiguous and any of the mentioned objects can be the pronoun's referent and (\textit{E. none of the above object}). Previous studies on how LLMs respond to multiple-choice questions have shown that LLMs prefer certain answer symbols, regardless of the actual content. In order to control for this effect, we shuffle the order of the options except for option D and E, resulting in a total of three different runs, and take the mean over runs. 

\begin{tcolorbox}[colback=white!5!white,colframe=black!75!black,title=Prompt template 2]

{\small \texttt{<USER> Consider the pronoun "it" in the following sentence: \\
Sentence: <sentence> \\
Which object does "it" refer to? \\
A. first object \\
B. second object  \\
C. mereological object  \\
D. any of the above objects  \\
E. none of the above objects \\
Answer: \\
<ASSISTANT>}
}

\end{tcolorbox}

See the excerpt below for an example prompt
and the LLM’s response.

\begin{quote}
    <USER> Consider the pronoun "it" in the following sentence: \\
Sentence: John put the card into the envelope and hid it \_\_. \\
Which object does "it" refer to? \\
A. the  card \\
B. the  envelope  \\
C. the  letter  \\
D. any of the above objects  \\
E. none of the above objects \\
Answer: \\
<ASSISTANT> The answer is A. the card. \\
In the sentence "John put the card into the envelope and hid it," the pronoun 'it' refers to the last object mentioned before it, which is 'the card.' This is a common usage of pronouns in English, where 'it' is often used to refer back to the nearest noun in a sentence.
\end{quote}

The answer choices are coded as follows.

\noindent
-- \texttt{Obj. 1}: The first object \\
-- \texttt{Obj. 2}: The second object \\
-- \texttt{Mereo.}: The mereological object \\
-- \texttt{Any.}: Any of the above objects \\
-- \texttt{None.}: None of the above objects \\

\subsection{Prompt with Explicit Mention of Ambiguity}

\subsubsection{Prompt 3: Open-ended Questions}

We add various hints of ambiguity to study whether LLMs' responses are affected. To see if the LLMs are aware of the mereological object. We directly ask the LLMs if the mereological objects can be the referent for \textit{it}. Responses are coded as either \texttt{Yes} or \texttt{No}.

\begin{tcolorbox}[colback=white!5!white,colframe=black!75!black,title=Prompt template 3]

\texttt{\small <USER> Consider the pronoun "it" in the following sentence. Can "it" refer to the \texttt{<mereological object>}? Provide the answer and explain. \\
Sentence: <sentence>} \\

\small \texttt{<ASSISTANT>}

\end{tcolorbox}

See the excerpt below for an example prompt
and the LLM’s response.

\begin{quote}
    <USER> Consider the referent of the pronoun 'it' in the following sentence: \\
Sentence: The shoemaker glued the leather to the sole and inspected it \_\_. \\
Can 'it' refer to the shoe? \\

<ASSISTANT>
Yes, "it" can refer to the shoe in this sentence. The context suggests that after the shoemaker has glued the leather to the sole, they then inspect the completed shoe or the work as a whole. Without additional context, the most natural and probable interpretation is that "it" refers to the shoe that has been recently made. However, depending on the broader context of the paragraph or the surrounding sentences, there could potentially be other referents for "it."

\end{quote}

\subsubsection{Prompt 4: Open-ended Questions}

Finally, we use an open question format to test whether LLMs capture the fact that the pronoun \textit{it} can refer to multiple referents. As shown in \texttt{prompt 2}, the model was asked to tell whether the pronoun \textit{it} is ambiguous and explain their answer. \texttt{<sentence>} is the sentence prefix.

\begin{tcolorbox}[colback=white!5!white,colframe=black!75!black,title=Prompt template 4]

\texttt{\small <USER> Is the pronoun "it" in the following sentence ambiguous? Provide the answer and explain. \\
Sentence: <sentence>} \\

\small \texttt{<ASSISTANT>}

\end{tcolorbox}

See the excerpt below for an example prompt
and the LLM’s response.

\begin{quote}
    <USER> Is the pronoun 'it' in the following sentence ambiguous? \\ 
    Provide the answer and explain. \\
Sentence: The railway worker attached the engine to the goods wagon and sent it \_\_.

<ASSISTANT> Yes, the pronoun 'it' in the given sentence is ambiguous. \\ 

The sentence "The railway worker attached the engine to the goods wagon and sent it \_\_" does not clearly specify whether 'it' refers to the 'engine' or the 'goods wagon'. Without additional context, there are two possible interpretations:

1. The railway worker attached the engine to the goods wagon and then sent (moved) the engine.
2. The railway worker attached the engine to the goods wagon and then sent (moved) the goods wagon.

The ambiguity arises because 'it' could be referring to either of the two nouns mentioned previously in the sentence without any clear indicators such as an article or modifying phrase. To remove this ambiguity, more context or specific language would be needed to clarify what exactly was sent.

\end{quote}

 LLMs responses are divided into: (1) \texttt{Unambiguous}, where the LLMs do not recognize the ambiguity of the pronoun, (2) \texttt{Ambiguous}, where the LLMs think that the two constituents are both possible to the referent, and (3) \texttt{Ambiguous + Mereo.}, where they also mention that the mereological object is also likely to be a referent.

\end{document}